\documentclass[11pt]{article}
\usepackage{amsmath}

\usepackage[preprint]{acl}

\usepackage{times}
\usepackage{latexsym}

\usepackage[T1]{fontenc}
\usepackage[utf8]{inputenc}

\usepackage{microtype}

\usepackage{inconsolata}

\usepackage{graphicx}
\graphicspath{{images/}}

\usepackage{tikz}
\usetikzlibrary{arrows.meta, positioning, fit, backgrounds}

\title{Evaluating the Impact of Task Granularity on Catastrophic Forgetting in Continual Learning}

\author{
Emre Alyamac\textsuperscript{1,2} \quad
Himanshu Janmeda\textsuperscript{1} \quad
Shashwat Krishna\textsuperscript{1,2} \quad
Yash Vijay\textsuperscript{1} \\
\\
\textsuperscript{1} College of Engineering \\
\textsuperscript{2} College of Natural Science \\
\\
\texttt{\{alyamace, singhjan, krish183, vijayyas\}@msu.edu}
}

\begin{document}
\maketitle

\begin{abstract}

Catastrophic forgetting, the abrupt loss of previously acquired knowledge upon learning new information, remains the central challenge in Continual Learning. This project investigates whether the order in which a model learns information affects how well it retains knowledge. Specifically, we ask: does learning general categories first (like "animals" vs "vehicles") before learning specific classes (like "dog" vs "cat") reduce forgetting compared to learning all classes at once?

We test three approaches on CIFAR-100: (1) Coarse-to-Fine: train on 2 super-classes, then expand to 10 specific sub-classes, (2) Fine-to-Coarse: train on 10 sub-classes, then group into 2 super-classes, and (3) Flat: train on all 10 classes from the start. We use Elastic Weight Consolidation (EWC) to prevent forgetting during transitions. Our hypothesis is that learning general patterns first creates a stable foundation that helps the model retain knowledge when learning more detailed distinctions. We evaluate using standard metrics (accuracy, precision, recall, F1) plus continual learning metrics like backward transfer and forgetting rates. This work could inform how we design learning sequences for real-world systems that need to learn incrementally.

Data and code: \url{https://github.com/burning-phoenix/ResNet-CL}
\end{abstract}

\section{Introduction}
Machine learning systems increasingly need to learn from data that arrives over time, rather than all at once. In these scenarios, an important question is: at what level of detail should we teach the model to recognize things?
In many real-world applications, data naturally has hierarchical structure. For example, a medical diagnosis system might first need to tell the difference between "Healthy" and "Sick" patients before learning to identify specific diseases. However, most continual learning research treats all classes as equally detailed and independent.
We investigate whether learning in a hierarchical order affects how well a model remembers what it learned. Our main hypothesis is: learning general categories first (Coarse-to-Fine) will help the model remember better than learning specific categories first (Fine-to-Coarse) or learning everything at the same level of detail (Flat).
We believe this works because learning coarse categories first forces the model to focus on broad, general features. These general features create a stable foundation. Later, when the model learns finer details like specific animal types, it can build on this foundation without forgetting the basics.

\section{Related Works}
The problem of "catastrophic forgetting" was first identified by McCloskey \& Cohen (1989). It describes how neural networks tend to forget what they learned previously when trained on new data, the weights get overwritten.
One popular solution is Elastic Weight Consolidation (EWC) by Kirkpatrick et al. (2017). EWC works by identifying which weights are most important for previous tasks and penalizing changes to those weights during new learning.
Other research has shown that hierarchical relationships between tasks can help with learning. For instance, Yosinski et al. (2014) found that features learned for one task transfer better to similar tasks. However, most continual learning experiments use datasets where tasks don't have natural hierarchical relationships (like Split-MNIST).
Our project is different because we specifically test how the order of learning (from general to specific vs. specific to general) affects forgetting in a dataset with clear hierarchical structure.

\section{Experimental Setup}

\begin{figure*}[t]
\centering
\begin{tikzpicture}[
    box/.style={draw, rounded corners, minimum width=2.2cm, minimum height=0.9cm, align=center, font=\small},
    activebox/.style={box, fill=blue!20, draw=blue!60},
    frozenbox/.style={box, fill=gray!15, draw=gray!40, text=gray!60},
    backbonebox/.style={box, fill=orange!20, draw=orange!60, minimum width=2.2cm},
    classbox/.style={draw=gray!50, rounded corners, fill=gray!8, font=\tiny, align=center, minimum width=2.2cm, inner sep=3pt},
    arrow/.style={->, thick, >=stealth},
    ewcarrow/.style={->, thick, red, dashed, >=stealth},
    waggarrow/.style={->, thick, blue!60, dashed, >=stealth},
    node distance=1.0cm
]

%% ─── COARSE TO FINE (left half) ───────────────────────────────────────────

\node[font=\bfseries\large] (ctf_title) at (2.8, 0.3) {Coarse $\to$ Fine};

% Task labels
\node[font=\small\itshape] (ctf_t1_label) at (0.8, -0.4) {Task 1 (Subset A)};
\node[font=\small\itshape] (ctf_t2_label) at (4.8, -0.4) {Task 2 (Subset B)};

% Transition arrow
\draw[->, thick, gray] (ctf_t1_label.east) -- node[above, font=\tiny\itshape] {transition} (ctf_t2_label.west);

% Task 1 boxes
\node[activebox]   (ctf_t1_coarse)   at (0.8, -1.6)  {Coarse Head\\{\tiny Linear(512,2)}\\{\tiny\color{blue!70}ACTIVE}};
\node[frozenbox]   (ctf_t1_fine)     at (0.8, -3.2)  {Fine Head\\{\tiny Linear(512,10)}\\{\tiny frozen}};
\node[backbonebox] (ctf_t1_backbone) at (0.8, -4.8)  {Shared Backbone\\{\tiny ResNet-18/MLP}};
\node[box]         (ctf_t1_input)    at (0.8, -6.2)  {Input Image\\{\tiny Subset A}};

% Task 1 arrows
\draw[arrow] (ctf_t1_input.north)    -- (ctf_t1_backbone.south);
\draw[arrow] (ctf_t1_backbone.north) -- (ctf_t1_fine.south);
\draw[arrow] (ctf_t1_fine.north)     -- (ctf_t1_coarse.south);

% Task 1 class box
\node[classbox] (ctf_t1_classes) at (0.8, -7.4) {vehicles\_1\\reptiles};

% Task 2 boxes
\node[frozenbox]   (ctf_t2_coarse)   at (4.8, -1.6)  {Coarse Head\\{\tiny Linear(512,2)}\\{\tiny frozen}};
\node[activebox]   (ctf_t2_fine)     at (4.8, -3.2)  {Fine Head\\{\tiny Linear(512,10)}\\{\tiny\color{blue!70}ACTIVE}};
\node[backbonebox] (ctf_t2_backbone) at (4.8, -4.8)  {Shared Backbone\\{\tiny ResNet-18/MLP}};
\node[box]         (ctf_t2_input)    at (4.8, -6.2)  {Input Image\\{\tiny Subset B}};

% Task 2 arrows
\draw[arrow] (ctf_t2_input.north)    -- (ctf_t2_backbone.south);
\draw[arrow] (ctf_t2_backbone.north) -- (ctf_t2_fine.south);
\draw[arrow] (ctf_t2_fine.north)     -- (ctf_t2_coarse.south);

% Task 2 class box
\node[classbox] (ctf_t2_classes) at (4.8, -7.4) {bicycle, bus,\\motorcycle,\\pickup\_truck, train,\\crocodile, dinosaur,\\lizard, snake, turtle};

% EWC arrow
\draw[ewcarrow] (ctf_t1_backbone.east) --
    node[above, font=\tiny, red, yshift=3pt] {EWC protects $\theta^*$}
    (ctf_t2_backbone.west);

%% ─── DIVIDER ───────────────────────────────────────────────────────────────

\draw[gray!40, dashed, thick] (6.8, 0.6) -- (6.8, -9.2);

%% ─── FINE TO COARSE (right half) ──────────────────────────────────────────

\node[font=\bfseries\large] (ftc_title) at (10.4, 0.3) {Fine $\to$ Coarse};

% Task labels
\node[font=\small\itshape] (ftc_t1_label) at (8.2, -0.4) {Task 1 (Subset A)};
\node[font=\small\itshape] (ftc_t2_label) at (12.2, -0.4) {Task 2 (Subset B)};

% Transition arrow
\draw[->, thick, gray] (ftc_t1_label.east) -- node[above, font=\tiny\itshape] {transition} (ftc_t2_label.west);

% Task 1 boxes
\node[frozenbox]   (ftc_t1_coarse)   at (8.2, -1.6)  {Coarse Head\\{\tiny Linear(512,2)}\\{\tiny frozen}};
\node[activebox]   (ftc_t1_fine)     at (8.2, -3.2)  {Fine Head\\{\tiny Linear(512,10)}\\{\tiny\color{blue!70}ACTIVE}};
\node[backbonebox] (ftc_t1_backbone) at (8.2, -4.8)  {Shared Backbone\\{\tiny ResNet-18/MLP}};
\node[box]         (ftc_t1_input)    at (8.2, -6.2)  {Input Image\\{\tiny Subset A}};

% Task 1 arrows
\draw[arrow] (ftc_t1_input.north)    -- (ftc_t1_backbone.south);
\draw[arrow] (ftc_t1_backbone.north) -- (ftc_t1_fine.south);
\draw[arrow] (ftc_t1_fine.north)     -- (ftc_t1_coarse.south);

% Task 1 class box
\node[classbox] (ftc_t1_classes) at (8.2, -7.4) {bicycle, bus,\\motorcycle,\\pickup\_truck, train,\\crocodile, dinosaur,\\lizard, snake, turtle};

% Task 2 boxes
\node[activebox]   (ftc_t2_coarse)   at (12.2, -1.6)  {Coarse Head\\{\tiny Linear(512,2)}\\{\tiny\color{blue!70}ACTIVE}\\{\tiny(aggregated init)}};
\node[frozenbox]   (ftc_t2_fine)     at (12.2, -3.2)  {Fine Head\\{\tiny Linear(512,10)}\\{\tiny frozen}};
\node[backbonebox] (ftc_t2_backbone) at (12.2, -4.8)  {Shared Backbone\\{\tiny ResNet-18/MLP}};
\node[box]         (ftc_t2_input)    at (12.2, -6.2)  {Input Image\\{\tiny Subset B}};

% Task 2 arrows
\draw[arrow] (ftc_t2_input.north)    -- (ftc_t2_backbone.south);
\draw[arrow] (ftc_t2_backbone.north) -- (ftc_t2_fine.south);
\draw[arrow] (ftc_t2_fine.north)     -- (ftc_t2_coarse.south);

% Task 2 class box
\node[classbox] (ftc_t2_classes) at (12.2, -7.4) {vehicles\_1\\reptiles};

% EWC arrow
\draw[ewcarrow] (ftc_t1_backbone.east) --
    node[above, font=\tiny, red, yshift=3pt] {EWC protects $\theta^*$}
    (ftc_t2_backbone.west);

%  arrow — from fine head T1 north, arcing to coarse head T2 west
\draw[waggarrow] (ftc_t1_fine.north)
    to[out=90, in=180, looseness=0.8]
    node[above right, font=\tiny, blue!70, yshift=2pt] {weight aggregation}
    (ftc_t2_coarse.west);

\end{tikzpicture}
\caption{Illustration of the coarse-to-fine (left) and fine-to-coarse (right)
learning trajectories. The shared backbone is updated during each task while
the inactive head is frozen. The red dashed arrow indicates EWC protection of
backbone weights between tasks. The blue dashed arrow in fine-to-coarse
indicates weight aggregation from the fine head to initialize the coarse head
at the transition point.}
\label{fig:trajectories}
\end{figure*}
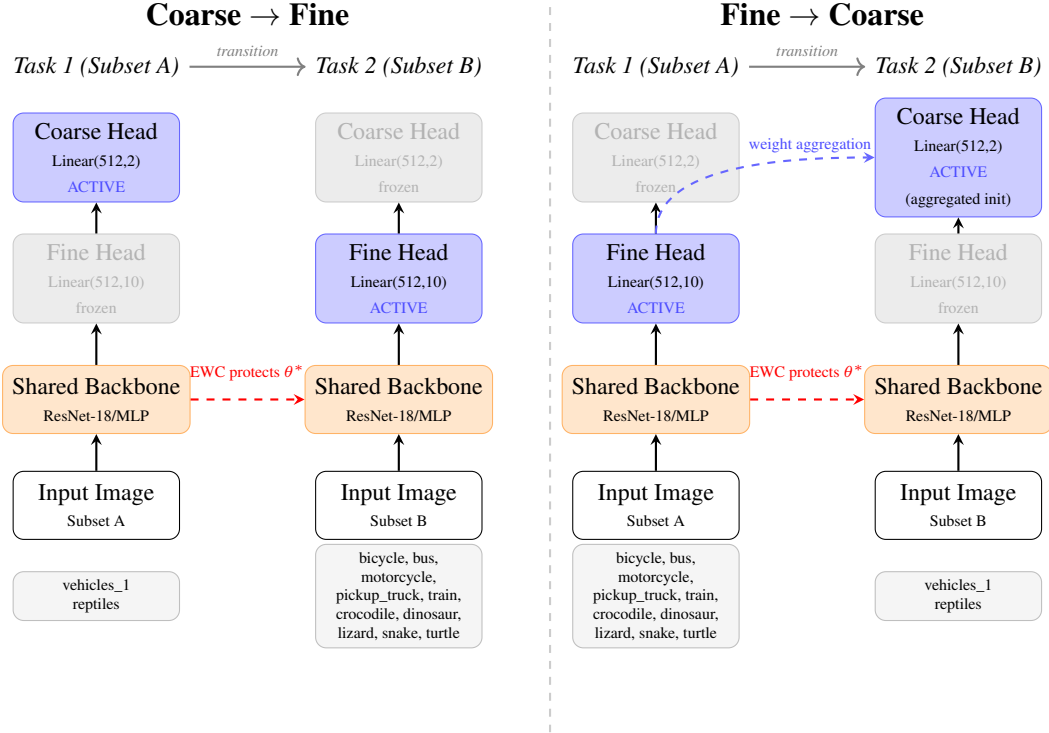

\begin{table*}[t]
\centering
\begin{tabular}{llp{6cm}cc}
\hline
\textbf{Superclass} & \textbf{Coarse Label} & \textbf{Fine Classes} & \textbf{Train Images} & \textbf{Test Images} \\
\hline
vehicles\_1 & 0 & bicycle, bus, motorcycle, pickup truck, train & 2,500 & 500 \\
reptiles    & 1 & crocodile, dinosaur, lizard, snake, turtle    & 2,500 & 500 \\
\hline
\textbf{Total} & & & \textbf{5,000} & \textbf{1,000} \\
\hline
\end{tabular}
\caption{The two CIFAR-100 superclasses used in our experiments. Each superclass contains 5 fine classes with 500 training and 100 test images per fine class.}
\label{tab:classes}
\end{table*}

We evaluate on CIFAR-100, a standard image classification benchmark consisting of 60,000 32×32 color images spanning 100 fine-grained classes organized into 20 superclasses. We restrict our experiments to two superclasses that provide a visually distinct hierarchy: vehicles\_1 and reptiles. The composition of the two superclasses are as in \autoref{tab:classes} This gives a total of 10 fine classes and 2 coarse classes, providing a clean testbed for studying granularity effects in continual learning.

\textbf{Task split:} Standard continual learning benchmarks split tasks by class, but allow the model to see the same images across runs, which violates the non-stationarity requirement of true sequential learning. To address this, we partition the training data at the instance level. For each of the 10 fine classes, the 500 available training images are divided into two deterministic, disjoint subsets of 250 images each: subset $A$, used exclusively for Task 1, and subset $B$, used exclusively for Task 2. The full set $A \cup B$  is used for the non-sequential flat baseline. This guarantees the model never sees the same image twice across tasks. The test set is kept intact with 100 images per fine class, giving 1,000 test images total, and is shared across all evaluation checkpoints.

\textbf{Preprocessing and augmentation:} All training images are augmented with random horizontal flipping and random 32×32 crops with 4-pixel zero-padding. Both training and test images are normalized using CIFAR-100 channel-wise mean (0.5071, 0.4867, 0.4408) and standard deviation (0.2675, 0.2565, 0.2761). No augmentation is applied at test time.

\textbf{Trajectories:} We evaluate three learning trajectories. In the \textbf{Coarse-to-Fine} trajectory, the model first trains on the 2-class superclass task (vehicles\_1 vs. reptiles) using subset $A$, then trains on the 10-class fine-grained task using subset $B$. In the \textbf{Fine-to-Coarse trajectory}, the order is reversed: the model first trains on the 10 fine classes using subset $A$, then on the 2 superclasses using subset $B$. In the Flat baseline, the model trains on all 10 fine classes using the full training set $A \cup B$ in a single non-sequential pass, serving as an upper bound on fine-grained classification performance without any forgetting pressure. The trajectories are illustrated in \autoref{fig:trajectories}

\section{Model Architecture}
We evaluate two model architectures under all trajectory and regularization conditions: a deep convolutional network (ResNet-18) and a linear baseline (Multinomial Logistic Regression). Both follow the same multi-head design, where a shared feature extractor feeds into two independent classification heads — one for the coarse task (2 classes) and one for the fine-grained task (10 classes). For each task, only the active head is given the gradient while the latter is frozen.

\textbf{ResNet-18}. Our primary model is a ResNet-18 trained from scratch with no pretrained weights. The default fully-connected output layer is removed and replaced with two independent linear heads: a coarse head Linear(512, 2) and a fine head Linear(512, 10), where 512 is the dimensionality of ResNet-18's penultimate feature vector. During training, the shared backbone and the active head are updated jointly via SGD. The 512-dimensional feature vectors produced by the backbone are also used for representation analysis (t-SNE and silhouette scoring) at each training checkpoint.

\textbf{Multinomial Logistic Regression}. As a linear baseline, we use a multi-head logistic regression model. Each input image is flattened from 3×32×32 to a 3,072-dimensional vector and passed through a shared single linear layer with ReLU activation, producing a 256-dimensional feature vector. This is then routed to either the coarse head Linear(256, 2) or the fine head Linear(256, 10), depending on the active task. Both models output raw logits; the softmax is applied implicitly by PyTorch's cross-entropy loss during training.

\textbf{Head initialization}. For the Fine-to-Coarse trajectory, the coarse head can be initialized before Task 2 training begins. This is because, unlike Coarse-to-Fine, training the fine head in Task 1 gives valuable information about the weights for the coarse head (fine to coarse mappings are known). Thus, instead of random initialization, we initialize the coarse head weights by aggregating the fine head weights of each superclass's child classes:
$$
    w_s^{\text{coarse}} = \frac{1}{|D_s|} \sum_{k \in D_s} w_k^{\text{fine}}
$$
where $D_s$ is the set of fine class labels corresponding to coarse label $s$.

\section{Training}
All trajectories and conditions share the same optimizer, architecture, and hyperparameters. The only variables across runs are the task ordering (trajectory) and the regularization method (condition).
\subsection{Hyperparameters}
All runs use the hyperparameters listed in Table 2.
\begin{table}[h!]
\centering
\begin{tabular}{ll}
\hline
\textbf{Hyperparameter} & \textbf{Value} \\
\hline
Optimizer & SGD \\
Learning rate & 0.01 \\
Momentum & 0.9 \\
Weight decay & 0 \\
Batch size & 128 \\
Epochs per task & 50 \\
Fisher samples & 2,000 \\
Fisher variant & Empirical \\
\hline
EWC $\lambda$ grid & \{0.01, 0.1, 1.0, 10.0\} \\
L2 $\lambda$ grid & \{0.01, 0.1, 1.0\} \\
Random seeds & \{42\} \\
\hline
\end{tabular}
\caption{Shared hyperparameters across all experimental runs.}
\label{tab
:hyperparams}
\end{table}

\subsection{Regularization}
Each trajectory is evaluated under three conditions. In the unregularized condition, the model trains sequentially with no protection of Task 1 weights, serving as a lower bound on retention. In the L2 condition, a penalty term is added during Task 2 training that discourages the backbone weights from deviating from their Task 1 values $\theta^*$:
\begin{equation}
E_{\text{L2}}(\theta) = E(\theta) + \frac{\lambda}{2} \sum_{i} \left(\theta_i - \theta^*_i\right)^2
    \label{eq:l2}    
\end{equation}
In the EWC condition, the same penalty is applied but weighted by the Fisher Information Matrix $F$, which measures how important each weight was for Task 1 performance:
\begin{equation}
E_{\text{EWC}}(\theta) = E(\theta) + \frac{\lambda}{2} \sum_{i} F_i\left(\theta_i - \theta^*_i\right)^2
    \label{eq:ewc}
\end{equation}

The Fisher is computed empirically on Task 1 data after Task 1 training completes, processing one sample at a time to ensure per-sample gradient accuracy. Crucially, the L2 penalty is not implemented via the optimizer's weight decay parameter — weight decay penalizes deviation from zero, whereas both L2 and EWC here penalize deviation from $\theta^*$, the weights learned after Task 1.

\subsection{Experimental Matrix}

\begin{table}[h!]
\centering
\begin{tabular}{|l|c|c|c|}
\hline
\textbf{Trajectory} & \textbf{Unreg.} & \textbf{L2 ($3\lambda$)} & \textbf{EWC ($3\lambda$)} \\
\hline
Coarse-to-Fine & 1 & 3 & 3 \\
Fine-to-Coarse & 1 & 3 & 3 \\
Flat baseline  & 1 & — & — \\
\hline
\textbf{Total runs} & 3 & 6 & 6 \\
\textbf{With 3 seeds} & 9 & 18 & 18 \\
\hline
\end{tabular}
\caption{Experimental matrix. Each cell is one full training run. Total runs including seeds: 45 + 3 joint reference = 48.}
\label{tab:expmatrix}
\end{table}

\section{Evaluation}

\begin{table}[h]
  \centering
  \small
  \begin{tabular}{p{0.2\columnwidth}p{0.72\columnwidth}}
    \hline
    \textbf{Metric} & \textbf{Technical Definition} \\
    \hline
    Backward Transfer & Influence of learning $T_n$ on performance of $T_{n-1}$. \\
    Forward Transfer & Influence of learning $T_{n-1}$ on performance of $T_n$. \\
    Forgetting & Peak accuracy minus final accuracy on a task. \\
    Precision & $\text{TP}/(\text{TP}+\text{FP})$ \\
    Recall & $\text{TP}/(\text{TP}+\text{FN})$ \\
    F1 Score & $2\cdot\text{Precision}\cdot\text{Recall}/(\text{Precision}+\text{Recall})$ \\
    \hline
  \end{tabular}
  \caption{Evaluation metrics for hierarchical continual learning.}
  \label{tab:metrics}
\end{table}

\begin{table*}[!htbp]
\centering
\begin{tabular}{llp{0.8\columnwidth}}
\hline
\textbf{Metric} & \textbf{Formula} & \textbf{Interpretation} \\
\hline
Backward Transfer (BWT) & $R_{2,1} - R_{1,1}$ & Negative = forgetting. Positive = backward facilitation. \\
\hline
Forgetting & $R_{1,1} - R_{2, 1}$ & Positive number when forgetting occurs. $= -\text{BWT}$ \\
\hline
Forward Transfer (FWT) & $R_{1, 2} - \frac{1}{K}$ & How much Task 1 helps Task 2 zero-shot. $K$ = number of Task 2 classes. \\
\hline
\end{tabular}
\caption{Continual learning metrics derived from the accuracy matrix $R$. BWT and Forgetting are the primary metrics of interest.}
\label{tab:metricsdetailed}
\end{table*}

We evaluate each run at two checkpoints: after Task 1 training and after Task 2 training. Based on the results, we construct a $2\times2$ matrix, $R$. Here, 

\begin{itemize}
    \setlength{\itemsep}{0pt}
    \item $R_{1,1}$: accuracy on Task 1 test set right after Task 1 training
    \item $R_{1,2}$: accuracy on Task 2 test set right after Task 1 training (zero-shot, head is random)
    \item $R_{2,1}$: accuracy on Task 1 test set after Task 2 training (\textbf{retention})
    \item $R_{2,2}$: accuracy on Task 2 test set after Task 2 training
\end{itemize}

The most important entry is $R_{2, 1}$, which directly measures how much of the Task 1 knowledge the model retrained after Task 2 training.

To assess continual learning, we use the metrics given in \autoref{tab:metricsdetailed}. Note that although we have listed FWT as a metric, it has minimal analytic value in this experiment since there are only two tasks.

Apart from continual learning metrics, we report classification metrics like per-class precision, recall, and F1 score on the active task's test set, following the definitions in \autoref{tab:metrics}. We report macro-averaged F1 as a summary statistic.

\section{Result}

% Figure 1 — full width
\begin{figure*}[t]
    \centering
    \includegraphics[width=\textwidth]{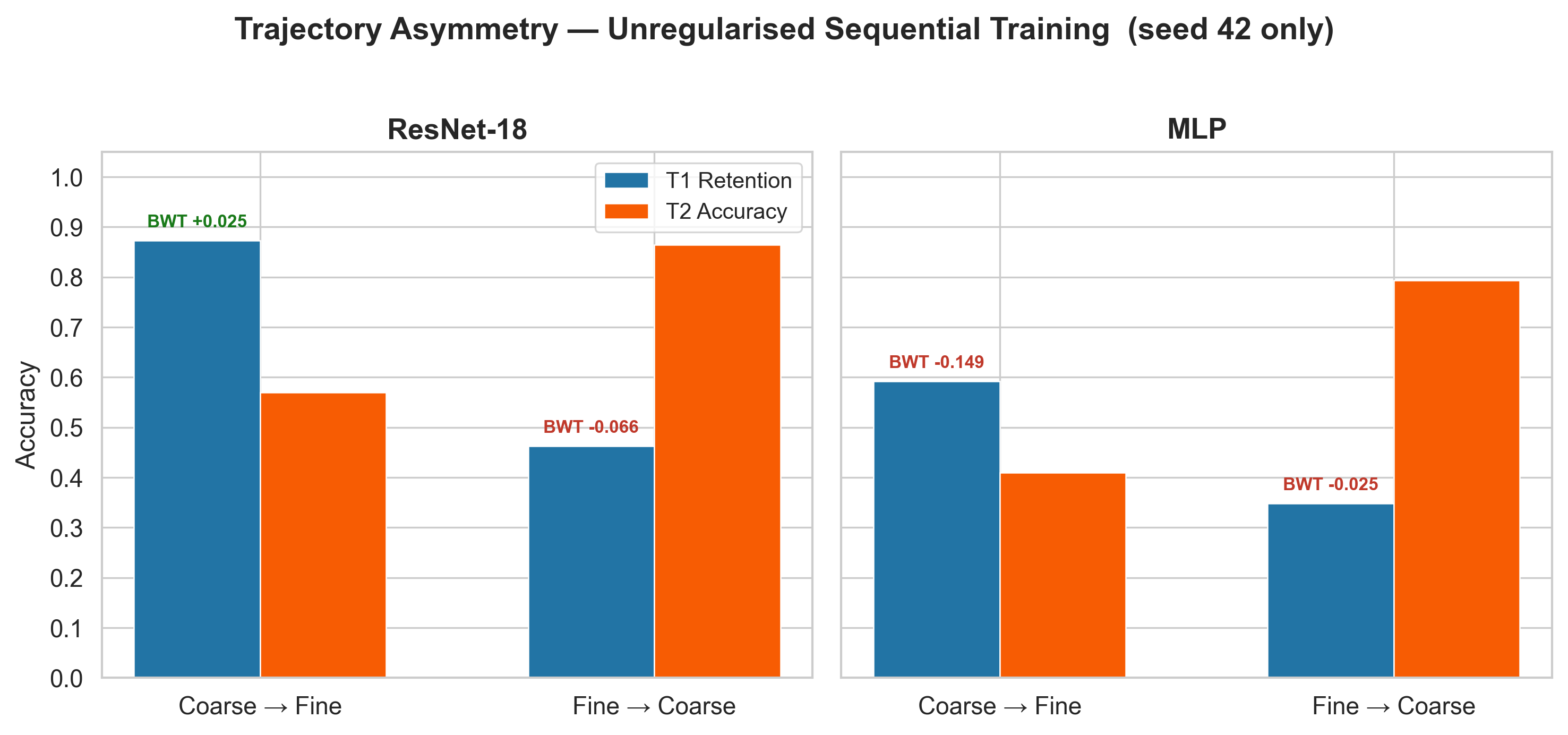}
    \caption{Trajectory asymmetry under unregularized sequential training (seed 42). 
    Blue bars show T1 retention ($R_{2,1}$) and orange bars show T2 accuracy ($R_{2,2}$). 
    BWT annotations indicate backward transfer for each trajectory. ResNet-18 shows 
    positive BWT for coarse-to-fine (+0.025) while MLP shows severe forgetting (-0.149).}
    \label{fig:asymmetry}
\end{figure*}

\textbf{Trajectory Asymmetry.} \autoref{fig:asymmetry} shows the T1 retention and T2 accuracy for both trajectories under unregularized sequential training. In the case of ResNet-18, coarse-to-fine has a BWT of $0.025$, indicating slight backward facilitation. Learning the fine-grained classes in Task 2, improved the coarse classification performance marginally, but does not hurt it. On the other hand the fine-to-coarse trajectory shows a BWT of $-0.066$, indicating forgetting of the fine-grained task after learning the coarser one. This is in line with our hypothesis that learning general (coarse, in this case) categories first provides a robust and stable foundation for the backbone of a model (in this case ResNet-18).

However, on the other hand, in the case of the MLP, we observe the opposite of our hypothesis. In the coarse-to-fine trajectory, MLP produces severe forgetting with a BWT of $-0.149$ while fine-to-coarse remains mild at $-0.025$. This revels that the trajectory asymmetry is not universal and we'll explore more in the \autoref{sec:discussion}

% Figure 3 — retention-plasticity tradeoff
\begin{figure*}[t]
    \centering
    \includegraphics[width=\textwidth]{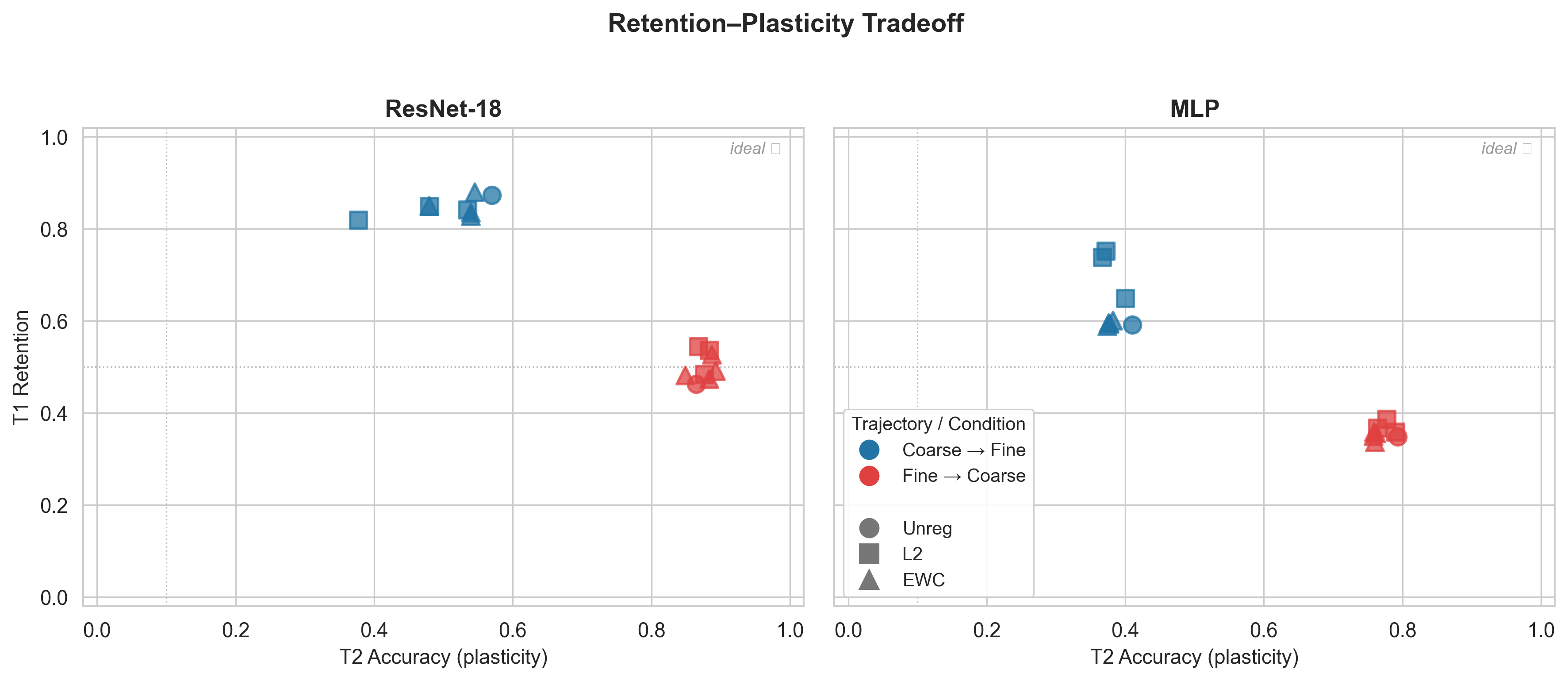}
    \caption{Retention-plasticity tradeoff across all trajectory and regularization 
    conditions. Each point is one run (seed 42). Color indicates trajectory, shape 
    indicates regularization condition. Ideal performance is top-right. ResNet-18 
    coarse-to-fine (blue) achieves the best retention at the cost of lower T2 accuracy.}
    \label{fig:tradeoff}
\end{figure*}

\textbf{Retention-Plasticity Tradeoff.} \autoref{fig:tradeoff} plots T1 retention against T2 accuracy across all trajectory and regularization conditions. An ideal continual 
learner would occupy the top-right corner — high retention and high plasticity simultaneously. In ResNet-18, the two trajectories are clearly separated. Coarse-to-fine sits in the top-left achieving high retention at the cost of lower T2 accuracy, while fine-to-coarse sits in the bottom right achieving high T2 accuracy but poor retention. This reveals an underlying pattern: the trajectory that remembers Task 1 better, learns Task 2 less effectively and vice-versa.

Notably, within each trajectory cluster, the three regularization conditions (Unreg, L2, EWC) produce very similar outcomes; the shapes are tightly clustered together. This suggests that for ResNet-18, the choice of trajectory dominates the choice of regularization method in determining the retention-plasticity operating point. 

The MLP shows a similar separation between trajectories but with weaker overall retention across both. Also, as seen in the case of ResNet-18, the shapes within each trajectory are clustered together indicating minimal effects from the choice of regularization strategies within a trajectory.

% Figure 4 — lambda sensitivity
\begin{figure*}[t]
    \centering
    \includegraphics[width=\textwidth]{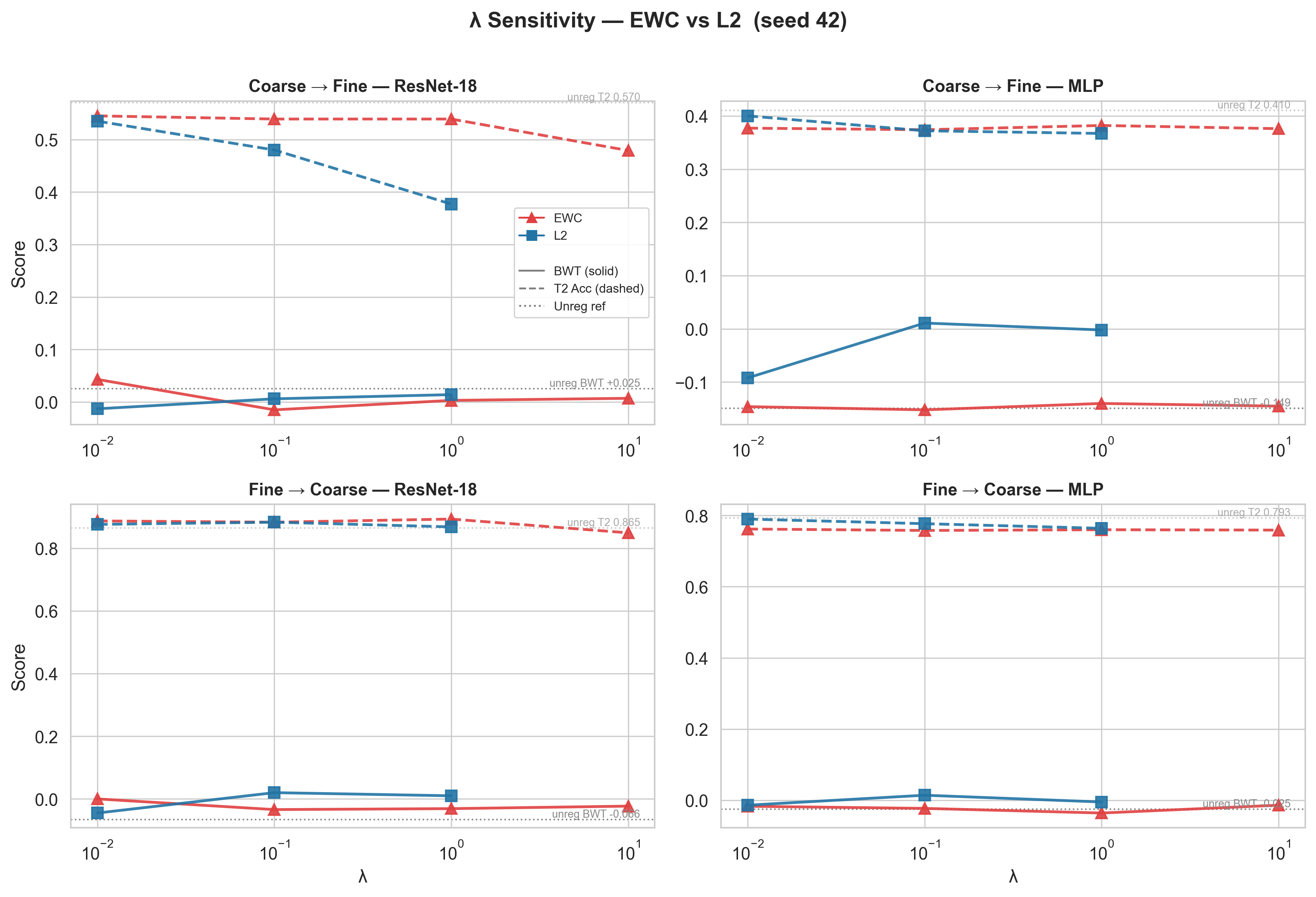}
    \caption{$\lambda$ sensitivity for EWC and L2 regularization across trajectories 
    and models (seed 42). Solid lines show BWT, dashed lines show T2 accuracy. 
    Dotted horizontal lines mark the unregularized reference. L2 BWT degrades at 
    high $\lambda$ in coarse-to-fine ResNet-18, while EWC remains stable.}
    \label{fig:lambda}
\end{figure*}

\textbf{Effect of Regularization.} \autoref{fig:lambda} shows how BWT and T2 accuracy vary with $\lambda$ for EWC and L2 across all trajectory and model combinations. Several patterns emerge.
\begin{itemize}
    \setlength{\itemsep}{0pt}
    \item First, for coarse-to-fine, L2 regularization shows a clear decrease in BWT as $\lambda$ increase. At high $\lambda$ values, L2 over-constrains the backbone and hurts T2 learning without meaningfully improving retention. EWC by contrast remains stable across the full $\lambda$ range, suggesting that Fisher-weighted protection is more robust to $\lambda$ choice than uniform L2 penalization which treats all weights as equally important regardless of their contribution to Task 1 performance.
    \item Second, for the fine-to-coarse trajectory in ResNet-18, neither L2 nor EWC produces meaningful changes in BWT or T2 accuracy across any $\lambda$ value; regularization is essentially ineffective in this trajectory. 
    \item Third, for the MLP in both trajectories, all curves are nearly flat regardless of $\lambda$, confirming that regularization is meaningless for the MLP.
\end{itemize}

\section{Discussion}
\label{sec:discussion}

\textbf{Trajectory asymmetry is real but capacity-dependent.} Our central finding is that the effect of task ordering on catastrophic forgetting is not universal; it is dependent on the model capacity and complexity. For ResNet-18, coarse-to-fine consistently outperforms fine-to-coarse in terms of T1 retention across 7 out of 8 regularization conditions, confirming our hypothesis. We attribute this to the hierarchical feature organization that deep residual networks develop naturally: coarse categorical boundaries are encoded in early layers while fine-grained discriminative features emerge in later layers. When the model first learns coarse structure, these early-layer representations remain stable during Task 2 training because fine-grained learning primarily updates downstream layers. The reverse trajectory disrupts this — fine-grained Task 1 training distributes discriminative features throughout the network, and the subsequent coarse Task 2 training overwrites them more broadly.

For the MLP, this explanation does not hold. A single hidden layer of 256 units cannot maintain separate representational hierarchies for coarse and fine structure simultaneously. When trained coarse-to-fine, the backbone must reorganize its limited representational capacity to accommodate the harder 10-class problem in Task 2, causing severe forgetting of the coarse boundary it learned. Fine-to-coarse avoids this because the easier Task 2 (2 classes) requires less reorganization of the features learned for the harder Task 1 (10 classes). This suggests that the benefit of hierarchical task ordering is contingent on the model having sufficient capacity to maintain layered abstractions.

\textbf{Regularization is trajectory-dependent.} A second key finding is that regularization effectiveness depends heavily on the trajectory. For coarse-to-fine ResNet-18, EWC provides stable BWT across all $\lambda$ values while L2 degrades at high $\lambda$, suggesting that Fisher-weighted protection is more appropriate when the backbone already has good coarse representations worth preserving. For fine-to-coarse ResNet-18, neither method provides meaningful benefit. The BWT curves are flat across all $\lambda$ values. We hypothesize this is because the Fisher Information computed after fine-grained Task 1 training is concentrated in later layers, leaving early layers under-protected during the coarse Task 2 update.

Unlike ResNet-18, for MLP, EWC and L2 are both entirely ineffective regardless of trajectory or $\lambda$. We hypothesize that this could perhaps be a result of the model's inability to capture significant features and details causing it to produce poor Fisher matrix values. 

\textbf{The retention-plasticity tradeoff.} \autoref{fig:tradeoff} reveals that the two trajectories do not simply differ in how much they forget. Coarse-to-fine trajectory sacrifices T2 accuracy for T1 retention, while fine-to-coarse sacrifices T1 retention for T2 accuracy. Neither trajectory dominates the other across both dimensions simultaneously. This suggests that in practical continual learning systems, the choice of task ordering should be guided by the dimension most relevant for the task: if retaining old knowledge is critical, coarse-to-fine is preferable; if learning new tasks well is the priority, fine-to-coarse may be more appropriate.

\section{Limitations}

Several limitations should be noted. First, all results reported here are from a single random seed (seed 42): multi-seed results are pending and conclusions may shift once variance is accounted for. Second, our experiment uses only two superclasses from CIFAR-100, which limits the generalizability of our findings to settings with more tasks or deeper hierarchies. Third, the instance-level A/B split reduces the training set to 250 images per class per task, which may disadvantage the MLP more than ResNet-18 given the former's limited capacity. Finally, the head weight aggregation used at the fine-to-coarse transition introduces an asymmetry between the two trajectories that may partially confound the comparison. Future work should control for this by also evaluating fine-to-coarse with random head initialization.

\section{Conclusion}

This work investigated whether the order in which a model learns hierarchical 
tasks affects catastrophic forgetting. Our results show that task ordering 
does matter, but not uniformly: for ResNet-18, learning coarse categories 
first reduces forgetting when transitioning to fine-grained classification, 
while the MLP shows the opposite pattern. This points to model capacity as 
a key factor: deep networks can maintain separate coarse and fine 
representations simultaneously, while shallow ones cannot. We also find that 
regularization method matters less than trajectory choice. EWC and L2 
produce similar retention-plasticity operating points, though EWC is more 
stable across $\lambda$ values. These findings suggest that in real-world 
continual learning systems, task ordering deserves as much attention as 
the choice of regularization algorithm. 

Future work should extend these findings to deeper hierarchies with more  than two superclasses, and evaluate across multiple random seeds to establish statistical reliability. It would also be valuable to control for the head initialization asymmetry between trajectories, and to explore whether more expressive regularization methods can close the retention-plasticity gap that we had in our experiment.

% Bibliography entries for the entire Anthology, followed by custom entries
\nocite{*}
\bibliography{custom}
\end{document}